\documentclass[twocolumn]{el-author}
\usepackage{algorithm,algpseudocode}
\usepackage{amsmath}

\newcommand{\ua}{\uparrow}
\newcommand{\nc}{\newcommand}
\nc{\da}{\downarrow} \nc{\hc}{\hat{c}} \nc{\hS}{\hat{S}}
\nc{\bra}{\langle} \nc{\ket}{\rangle} \nc{\eq}{equation (\ref}
\nc{\h}{\hat} \nc{\hT}{\h{T}}\nc{\be}{\begin{eqnarray}}
\nc{\ee}{\end{eqnarray}}\nc{\rd}{\textrm{d}}\nc{\e}{eqnarray}\nc{\hR}{\hat{R}}\nc{\Tr}{\mathrm{Tr}}
\nc{\tS}{\tilde{S}}\nc{\tr}{\mathrm{tr}}\nc{\8}{\infty}\nc{\lgs}{\bra\ua,\phi|}\nc{\rgs}{|\ua,\phi\ket}
\nc{\hU}{\hat{U}}\nc{\lfs}{\bra\phi|}\nc{\rfs}{|\phi\ket}\nc{\hZ}{\hat{Z}}\nc{\hd}{\hat{d}}\nc{\mD}{\mathcal{D}}
\nc{\bd}{\bar{d}}\nc{\bc}{\bar{c}}\nc{\mc}{\mathcal}\nc{\ea}{eqnarray}\nc{\mG}{\mathcal{G}}\nc{\bce}{\begin{center}}
\nc{\ece}{\end{center}}
\date{}

\begin{document}

\title{Two-Bit Networks for Deep Learning on Resource-Constrained Embedded Devices}

\author{Wenjia Meng, Zonghua Gu, Ming Zhang, Zhaohui Wu}

\abstract{With the rapid proliferation of Internet of Things and intelligent edge devices, there is an increasing need for implementing machine learning algorithms, including deep learning, on resource-constrained mobile embedded devices with limited memory and computation power. Typical large Convolutional Neural Networks (CNNs) need large amounts of memory and computational power, and cannot be deployed on embedded devices efficiently. We present Two-Bit Networks (TBNs) for model compression of CNNs with edge weights constrained to (-2, -1, 1, 2), which can be encoded with two bits. Our approach can reduce the memory usage and improve computational efficiency significantly while achieving good performance in terms of classification accuracy, thus representing a reasonable tradeoff between model size and performance.}

\maketitle

\section{Introduction}


Deep Neural Networks (DNNs) are widely used for a variety of machine learning tasks currently, including computer vision, natural language processing, and others. Convolutional Neural Networks (CNNs) are especially suitable for computer vision applications, including object recognition, classification, detection, image segmentation, etc. Smart mobile devices equipped with high-resolution cameras have opened the door to potential mobile computer vision applications. However, a typical CNN has millions of parameters and perform billions of arithmetic operations for each inference, e.g., AlexNet \cite{Krizhevsky12} has 61M parameters (249MB of memory with floating point weights) and performs 1.5B high-precision operations to classify one image. The enormous demand for memory storage and computation power hinders their deployment on resource-constrained embedded devices with limited memory and computation power. It is an active research topic to adapt DNNs/CNNs for deployment on embedded devices.

A number of authors have proposed techniques for reducing size of DNNs/CNNs, including model pruning \cite{Han15}, i.e., removing edges with small weight magnitudes, and \emph{weight compression}, i.e., reducing the precision of weights by using a small number of bits to edge weight and/or activation. These two approaches are orthogonal and can be applied separately or together. We focus on weight compression in this paper. The low bit-width translates to small memory footprint and high arithmetic computational efficiency. Courbariaux et al. presented \emph{BinaryConnect} \cite{Courbariaux15} for training a DNN with binary weights -1 or 1, and \emph{BinaryNet} \cite{Courbariaux16}for training a DNN with both binary weights and binary activations. Both BinaryConnect and BinaryNet can achieve good performance on small datasets such as MNIST, CIFAR-10 and SVHN, but performed worse than their full-precision counterparts by a wide margin on large-scale datasets like ImageNet. Rastegari et al.  \cite{Rastegari16} presented \emph{Binary Weight Networks} and \emph{XNOR-Net}, two efficient approximations to standard CNNs, which are shown to outperform BinaryConnect and BinaryNet by large margins on ImageNet. In Binary-WeightNetworks, the convolutional filters are approximated with binary values (-1,1); in XNOR-Networks, both filters and input to convolutional layers are binary. However, there is still a significant performance gap between these network models and their full-precision counterparts. Also, binarization of both activations and weights generally leads to dramatic performance degradation compared to binarization of weights only.

To strike a balance between model compression rate and model capacity, Li et al. \cite{Li16} presented \emph{Ternary Weight Networks} with weights constrained to (-1, 0, 1), each weight encoded with two bits. Compared with DNN models with binary weights, Ternary Weight Networks can achieve better performance due to increased weight precision. However, Ternary Weight Networks make use of only three values (-1, 0, 1) out of the four possible values that can be encoded with two bits.

In this paper, we propose Two-Bit Networks (TBNs) to further explore the tradeoff between model compression rate and model capacity. (We focus on CNNs in this paper, although our techniques can be adapted to apply to general DNNs, including Recurrent Neural Networks.) We constrain weights to four values (-2, -1, 1, 2), which can be encoded with two bits. Compared with existing weight compression methods, TBNs make more efficient use of weight bit-width to achieve higher model capacity and better performance. Arithmetic operations can be implemented with additions, subtractions, and shifts, which are very efficient and hardware-friendly.

We propose a training algorithm for DNNs based on Stochastic Gradient Descent. During each iteration, a set of real-valued weights are discretized into two-bit values, which are used by the following forward pass and backward pass. Then, the real-valued weights are updated with gradients computed by the backward pass. During inference, only the two-bit weights are used. Experimental results show that or method achieves better performance on ImageNet than other weight compression methods. 

\section{Two-Bits Networks}


In a CNN with $L$ layers, each layer $l \in \{ 1, \ldots, L \}$ performs a convolution $*$ on its input tensor $I^l \in \mathbb{R} ^{c^l \times w_{in}^l \times h_{in}^l }$ and each of its $K^l$ convolution filters $W^{l,k} \in \mathbb{R} ^{c^l \times w^l \times h^l }$, where $(c^l,w_{in}^l,h_{in}^l)$ and $(c^l,w^l,h^l)$ represents shape of the input tensor and the filter, respectively, including \textit{channels}, \textit{width}, and \textit{height}. Let $n^{l,k}=c^{l,k} \times w^{l,k} \times h^{l,k}$ denote the number of elements in $W^{l,k}$, and $W^{l,k}_i$ denote the $i^{th}$ element of $W^{l,k}$, with $i \in \{ 1, \ldots , n^{l,k} \}$. For brevity, we drop indexes $l$ and $k$ when they are unnecessary.

Each real-valued convolution filter $W$ is approximated with a binary filter $\tilde{W} \in \{ -2,-1,1,2 \} ^{c \times w \times h}$ and a scaling factor $ \alpha \in \mathbb{R} ^+$, so that $W \approx \alpha \tilde{W}$. A convolution operation can be approximated by:
\begin{align}
I*W \approx I*( \alpha \tilde{W})=( \alpha I) \oplus \tilde{W}
\label{approximateConvolution}
\end{align}
where $ \oplus $ denotes a convolution operation without multiplication.

Ideally, the TBN should mimic its full-precision counterpart closely, provided that quantization error between $W$ and its approximation $\alpha \tilde{W}$ is minimized. We seek to minimize the L2-norm of the quantization error for each convolution filter:
\begin{align}
\alpha ^*, \tilde{W} ^*= & \mathop{ \arg \min } \limits _{ \alpha , \tilde{W}} \parallel W- \alpha \tilde{W} \parallel _2^2 \nonumber \\
s.t. \qquad & \alpha > 0,\tilde{W} _i \in \{ -2,-1,1,2 \} , i \in \{ 1, \ldots ,n \}
\end{align}

The optimization can be divided into two steps. First, the real-valued weights are discretized to find the two-bit weights. Then, the optimal scaling factor is found to minimize the quantization error, given the two-bit weights. For simplicity, we adopt deterministic discretization:
\begin{align}
\tilde{W} _i=
    \begin{cases}
    -2 & \text{if } W_i<-1, \\
    -1 & \text{if } -1 \leq W_i \leq 0, \\
    1 & \text{if } 0< W_i \leq 1, \\
    2 & \text{if } W_i>1.
    \end{cases}
\label{discretization}
\end{align}

Substitute the two-bit weights (\ref{discretization}) into the expression for quantization error, the expression can be simplified into:
\begin{align}
J( \alpha ) & = \parallel W- \alpha \tilde{W} \parallel _2^2 \nonumber \\
& = \sum_{i:|W_i| \leq 1}{||W_i|- \alpha |^2} + \sum_{i:|W_i|>1}{||W_i|- 2 \alpha |^2} \nonumber \\
& = (|B_1|+4|B_2|) \alpha ^2 -2( \sum_{i \in B_1}{|W_i|} + 2 \sum_{i \in B_2}{|W_i|} ) \alpha +C
\end{align}
where $B_1= \{ 1 \leq i \leq n \big| |W_i| \leq 1 \}$, $B_2= \{ 1 \leq i \leq n \big| |W_i|>1 \}$, $C= \sum _{i=1}^n{W_i^2}$ is a constant independent of $ \alpha $, and $| W_i |$ denotes the magnitude of $W_i$.

Taking the derivative of $J( \alpha )$ w.r.t. $ \alpha $ and setting $ \frac{dJ( \alpha )}{d \alpha }$ to zero, we obtain the optimal scaling factor:
\begin{equation}
\alpha ^*= \frac{ \sum_{i \in B_1}{|W_i|} + 2 \sum_{i \in B_2}{|W_i|} }{|B_1|+4|B_2|}
\label{optimalScalingFactor}
\end{equation}

\section{Training Two-Bit Networks}


We describe details of the training algorithm for Two-Bit Networks based on Stochastic Gradient Descent (SGD). Algorithm \ref{alg1} shows the pseudocode for each training iteration. In order to keep track of tiny weight updates of each iteration, we adopt a similar trick as \cite{Courbariaux15} and maintain a set of real-valued convolution filters throughout the training process. First, approximate filters are computed from the real-valued filters for all convolutional layers (Lines \ref{trainQuantizationBegin}-\ref{trainQuantizationEnd}). Note that fully-connected layers can be treated as convolutional layers \cite{Shelhamer16}. The real-valued weights are discretized into two-bit weights (Line \ref{trainDiscretization}), and an optimal scaling factor is computed for each filter (Line \ref{trainScalingFactor}). Then, a forward pass is run on the network inputs (Line \ref{trainForwardPass}), followed by a backward pass, which back-propagates errors through the network to compute gradients w.r.t. the approximate filters (Line \ref{trainBackwardPass}).  Unlike conventional CNNs, the forward and backward passes use approximate filters instead of the real-valued filters. Finally, the real-valued filters are updated with the gradients (Line \ref{trainWeightUpdate}). During inference, only the two-bit filters and the optimal scaling factors are used.

\begin{algorithm}
\caption{SGD Training for Two-Bit Networks}
\label{alg1}
\begin{algorithmic}[1]
\Require
A minibatch of inputs $X$ and targets $Y$, current real-valued weights $W$ and learning rate $ \eta $
\Ensure
Updated real-valued weights $W^{new}$ and learning rate $ \eta ^{new}$
\For{$l=1\text{ to }L$} \label{trainQuantizationBegin}
\For{$k^{th}$ filter in $l^{th}$ layer}
    \State Compute two-bit filter $ \tilde{W} ^{l,k}$ from $W^{l,k}$ by (\ref{discretization}) \label{trainDiscretization}
    \State Compute optimal scaling factor ${ \alpha ^{l,k}}^*$ by (\ref{optimalScalingFactor}) \label{trainScalingFactor}
    \State Approximate filter $ \hat{W} ^{l,k}={ \alpha ^{l,k}}^* \tilde{W} ^{l,k}$
\EndFor
\EndFor \label{trainQuantizationEnd}
\State $\hat{Y}=\textbf{TwoBitForward}(X, \{ \tilde{W} ^{l,k} \} , \{ { \alpha ^{l,k}} ^* \} )$ // Convolutions by (\ref{approximateConvolution}) \label{trainForwardPass}
\State $ \{ \frac{ \partial C}{\partial \hat{W} ^{l,k}} \} =\textbf{TwoBitBackward}( \frac{ \partial C}{ \partial \hat{Y}} , \{ \hat{W} ^{l,k} \})$ \label{trainBackwardPass}
\State $W^{new}=\textbf{UpdateParameters}(W, \{ \frac{ \partial C}{\partial \hat{W} ^{l,k}} \} , \eta )$ \label{trainWeightUpdate}
\State Update learning rate $ \eta ^{new}$ according to any learning rate scheduling function
\end{algorithmic}
\end{algorithm}

\section{Experiments}

We use the well-known ImageNet dataset (ILSVRC2012) to evaluate performance of Two-Bit Networks. ImageNet is a computer vision benchmark dataset with a large number of labeled images,  divided into a training set and a validation set. The training set consists of $\sim$1.2M images from 1K categories, including animals, plants, and other common items; the validation set contains 50K images.

We use Deep Residual Networks (DRNs) \cite{He15} as the CNN architecture in our experiments, which achieved state-of-the-art performance on ImageNet. Compared with other CNN architectures, DRNs can have a large number of convolutional layers (from 18 to 152), and a shortcut connection that performs linear projection exists alongside each group of two consecutive convolutional layers, in order to reformulate the layers as learning functions with reference to the layer inputs.  For simplicity, we adopt ResNet-18, which has 18 convolutional layers and is the smallest model presented in their paper.

The experiments are conducted with Torch7 \cite{Collobert11} on an NVIDIA Titan X. At training time, images are randomly cropped with 224$\times$224 windows. We run the training algorithm for 58 epochs with batch size of 256. We use SGD with momentum of 0.9 to update parameters and batch normalization \cite{Ioffe15} to speed up convergence. The weight decay is set to 0.0001. The learning rate starts at 0.1 and is divided by 10 at epoches 30, 40, and 50. At inference time, we use the 224$\times$224 center crops for forward propagation.

\begin{figure}
\centering{\includegraphics[width=60mm]{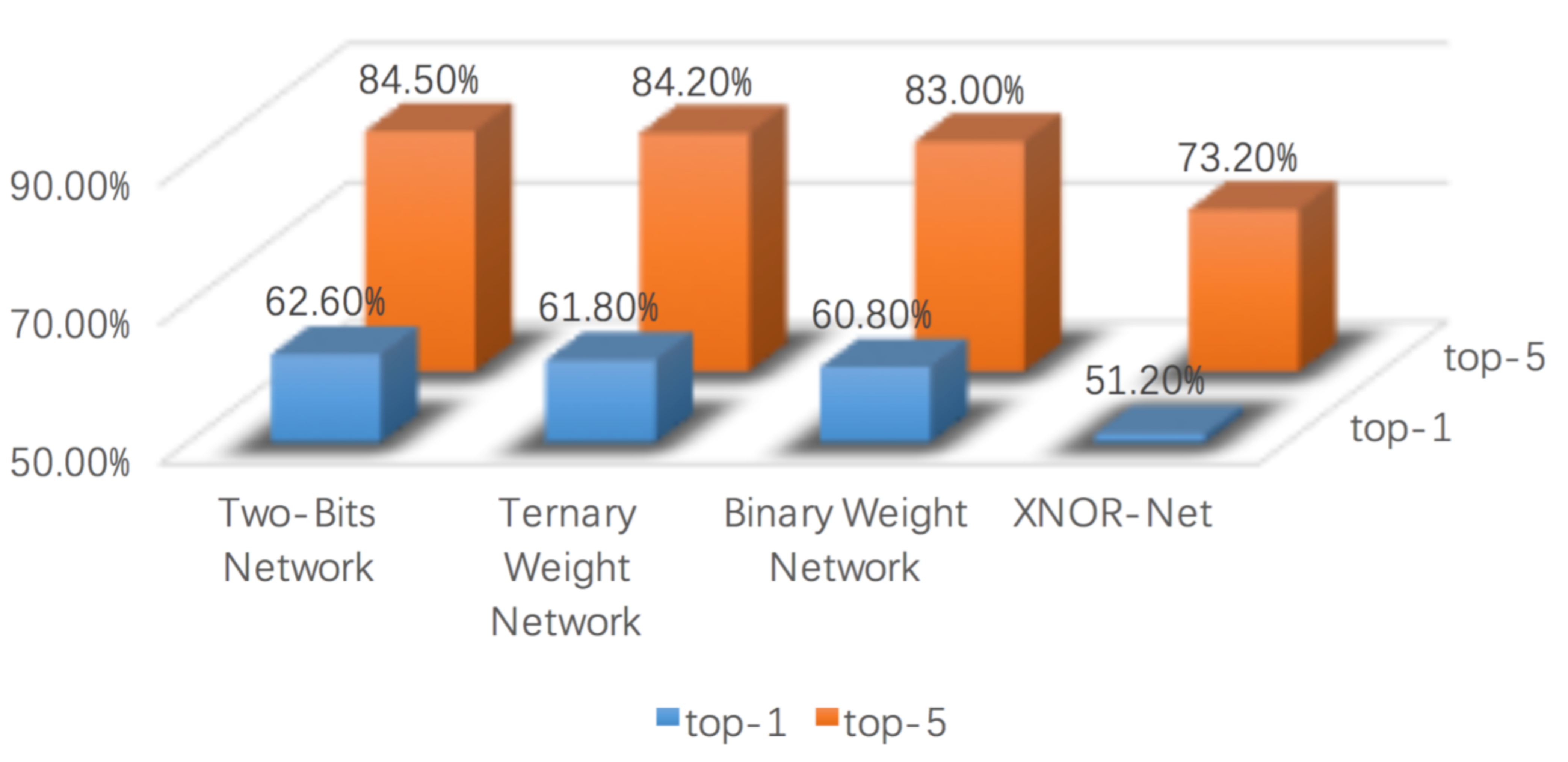}}
\caption{Validation accuracy comparison of the Two-Bit Network method and state-of-the-art weight compression methods on the ImageNet dataset}
\label{accuracies}
\end{figure}

We compare our method with state-of-the-art weight compression methods, including Ternary Weight Network \cite{Li16}, Binary Weight Network and XNOR-Net \cite{Rastegari16}. Fig. \ref{accuracies} shows performance results (classification accuracy) on ImageNet. Our method outperforms the other weight compression methods, with top-5 accuracy of 84.5\% and top-1 accuracy of 62.6\%. We attribute the improved performance to the increased model capacity due to more efficient use of the two-bit representation for weights. 

\begin{figure}
\centering{\includegraphics[width=60mm]{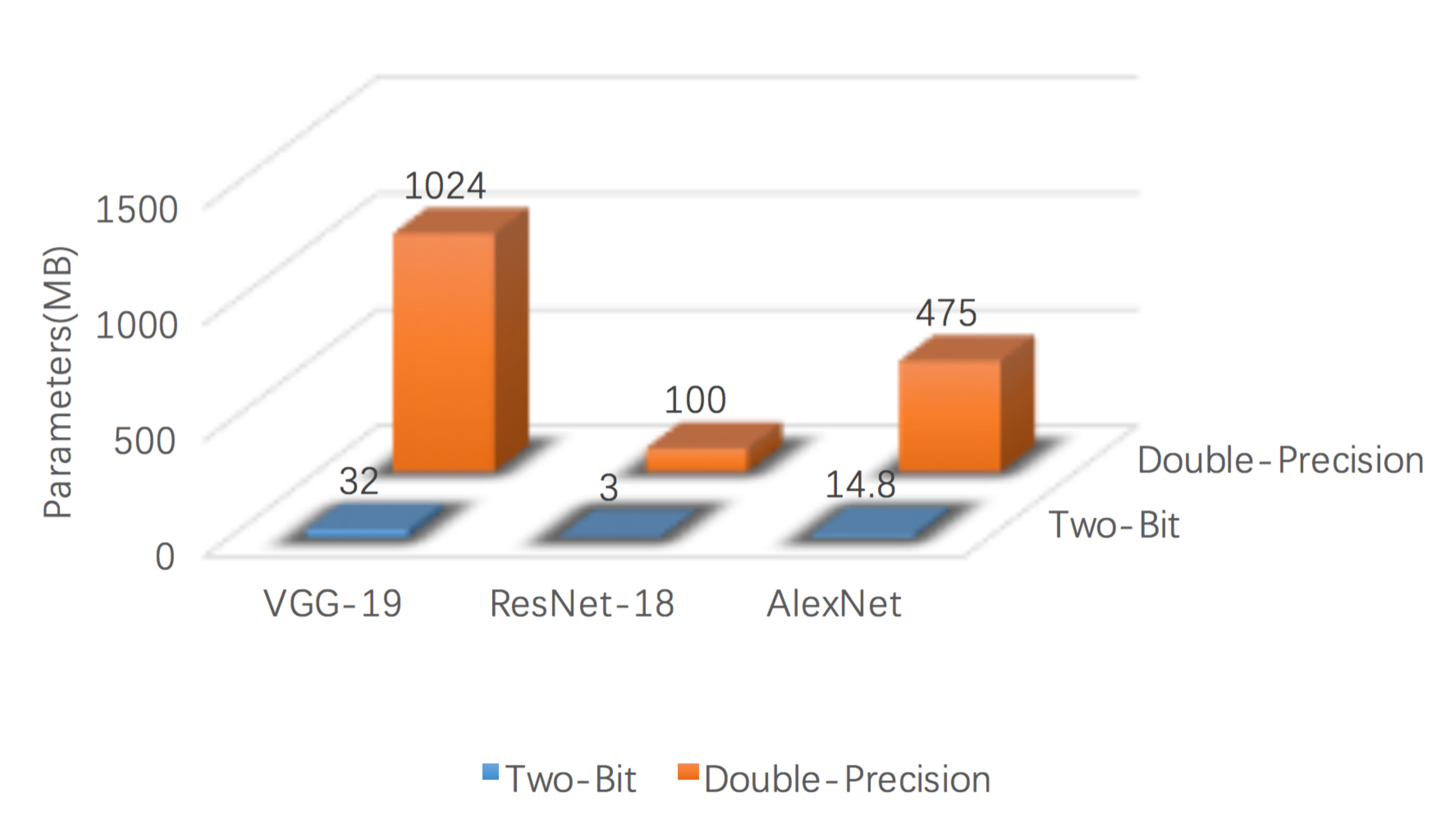}}
\caption{Memory size comparisons between Two-Bits Networks and double-precision networks for different CNN architectures.}\label{memorysize}
\end{figure}

Fig. \ref{memorysize} shows memory size requirement of Two-Bit Networks compared to double-precision floating point representation for three different architectures(AlexNet, ResNet-18 and VGG-19). The dramatic reduction in memory size requirement makes Two-Bit Networks suitable for deployment on embedded devices.

\section{Conclusion}
We have presented Two-Bit Networks for model compression of CNNs, which achieves a good tradeoff between model size and performance compared with state-of-the-art weight compression methods. Compared to the recent work on binary weights and/or activations, our method achieves higher model capacity and better performance with slightly larger memory size requirement. 
\vskip3pt
\ack{This work is partially supported by NSFC Grant \#61672454. The Titan-X GPU used for this research was donated by the NVIDIA Corporation.}

\vskip5pt

\noindent Wenjia Meng, Zonghua Gu, Ming Zhang and Zhaohui Wu (\textit{College of Computer Science, Zhejiang University, Hangzhou, China, 310027})
\vskip3pt

\noindent E-mail: zgu@zju.edu.cn


\begin{thebibliography}{}\small

\bibitem{Krizhevsky12}
Krizhevsky, A., Sutskever, I., Hinton, G.E., "Imagenet classification with deep convolutional neural networks,", Advances in neural information processing systems. (2012) 1097¨C1105.

\bibitem{Rastegari16}
M. Rastegari, V. Ordonez, J. Redmon, and A. Farhadi, "XNOR-Net: ImageNet Classification Using Binary Convolutional Neural Networks," CoRR, vol. abs/1603.0, 2016.

\bibitem{Courbariaux15}
M. Courbariaux, Y. Bengio, and J.-P. David, "BinaryConnect: Training Deep Neural Networks with binary weights during propagations," in Advances in Neural Information Processing Systems 28, 2015, pp. 3123-3131.

\bibitem{Courbariaux16}
M. Courbariaux and Y. Bengio, "Binary Neural Networks: Training Deep Neural Networks with Weights and Activations Constrained to +1 or -1," CoRR, vol. abs/1602.0, 2016.

\bibitem{Li16}
F. Li and B. Liu, "Ternary Weight Networks," CoRR, vol. abs/1605.04711, 2016.

\bibitem{Han15}
S. Han, J. Pool, J. Tran, and W. Dally, "Learning both Weights and Connections for Efficient Neural Network," in Advances in Neural Information Processing Systems 28, 2015, pp. 1135-1143.



\bibitem{Shelhamer16}
J. Long, E. Shelhamer, and T. Darrell, "Fully Convolutional Networks for Semantic Segmentation," in The IEEE Conference on Computer Vision and Pattern Recognition (CVPR), 2015.

\bibitem{He15}
K. He, X. Zhang, S. Ren, and J. Sun, "Deep Residual Learning for Image Recognition," CoRR, vol. abs/1512.03385, 2015.

\bibitem{Collobert11}
R. Collobert, K. Kavukcuoglu, and C. Farabet, "Torch7: A matlab-like environment for machine learning," in BigLearn, NIPS Workshop, 2011, no. EPFL-CONF-192376.

\bibitem{Ioffe15}
S. Ioffe and C. Szegedy, "Batch Normalization: Accelerating Deep Network Training by Reducing Internal Covariate Shift," Proc. 32nd Int. Conf. Mach. Learn., pp. 448-456, 2015.











\end{thebibliography}
\end{document}